\documentclass[review]{elsarticle}
\usepackage{graphicx}
\usepackage{subfig}
\usepackage{epsfig}
\usepackage{multicol}
\usepackage{float}
\usepackage{bm}
\usepackage{amsmath}
\usepackage{threeparttable}
\usepackage[table,xcdraw]{xcolor}
\usepackage{multirow}
\usepackage{amssymb}

\journal{Neurocomputing}

\begin{document}

\begin{frontmatter}

\title{Fully Convolutional Measurement Network for Compressive Sensing Image Reconstruction}

\author{Jiang Du}
%\ead{jiangdu@ieee.org}

\author{Xuemei Xie\corref{cor1}}
\ead{xmxie@mail.xidian.edu.cn}

\author{Chenye Wang}

\author{Guangming Shi}

\author{Xun Xu}

\author{Yuxiang Wang}

\cortext[cor1]{Corresponding author}

\address{School of Artificial
Intelligence, Xidian University, Xi'an, Shaanxi 710071, PR China}

\begin{abstract}
Recently, deep learning methods have made a significant improvement in compressive sensing image reconstruction task. In the existing methods, the scene is measured block by block due to the high computational complexity. This results in block-effect of the recovered images. In this paper, we propose a fully convolutional measurement network, where the scene is measured as a whole. The proposed method powerfully removes the block-effect since the structure information of scene images is preserved. To make the measure more flexible, the measurement and the recovery parts are jointly trained. From the experiments, it is shown that the results by the proposed method outperforms those by the existing methods in PSNR, SSIM, and visual effect.
\end{abstract}

\begin{keyword}
compressive sensing, full image measurement, block-effect, fully convolutional measurement network, convolutional neural network.
\end{keyword}

\end{frontmatter}

%\linenumbers
%% main text
\section{Introduction}\label{Introduction}
Compressive sensing (CS) theory \cite{baraniuk2010model,baraniuk2011more,candes2007sparsity,haupt2010toeplitz} is able to acquire measurements of signals at sub-Nyquist rates and recover signals with high probability when the signals are sparse in a certain domain. Greedy algorithms \cite{needell2010cosamp,babacan2010bayesian}, convex optimization algorithms \cite{candes2005decoding,figueiredo2007gradient}, and iterative algorithms \cite{donoho2009message,daubechies2004iterative} have been used for recovering images in conventional CS. However, almost all these methods recover images by solving an optimization problem, which is time-consuming.
In order to reduce the computational complexity in the reconstruction stage, convolutional neural networks (CNNs) are applied to replace the optimization process. CNN-based methods \cite{mousavi2015deep,kulkarni2016reconnet,mousavi2017learning,mousavi2017deepcodec,xie2017adaptive} use big data \cite{stevens2017} to train the networks that speed up the reconstruction stage. Mousavi, Patel, and Baraniuk \cite{mousavi2015deep} firstly propose deep learning approach to solve the CS recovery problem. They use stacked denoising autoencoders (SDA) to recover signals from undersampled measurements. ReconNet \cite{kulkarni2016reconnet} and DeepInverse \cite{mousavi2017learning} introduce CNNs to the reconstruction problem, where the random Gaussian measurement matrix is used to generate the measurements. Instead, the methods \cite{mousavi2017deepcodec,xie2017adaptive} using adaptive measurement learn a transformation from signals to the measurements. This mechanism allows measurements to retain more information from images. The methods mentioned above divide an image into blocks, which breaks the structure information of the image. This will cause the block effect in the reconstructed image.

In this paper, we propose a fully convolutional measurement network for CS image reconstruction. Instead of block-wise methods, a convolutional layer is applied to obtain the measurement from a full image, which keeps the integrity of structure information of the original image. Furthermore, motivated by the residual learning proposed by ResNet \cite{he2016deep}, we apply residual connection block (Resblock) in the proposed network for improvement. Experimental results show that the proposed method outperforms the state-of-the-art method 1$\sim$2dB in PSNR at different measurement rates.

The organization of this paper is as follows. The related works using deep learning methods for the CS reconstruction problem are introduced in Section \ref{RelatedWork}. Section \ref{section3} presents the proposed fully convolutional measurement network. Section \ref{section4} shows experimental results of the proposed method and the previous works. The conclusion of this paper is drawn in Section \ref{conclusion}.

\section{Related Work}\label{RelatedWork}
\begin{figure}[h!]
	\centering
	\includegraphics[width=\linewidth]{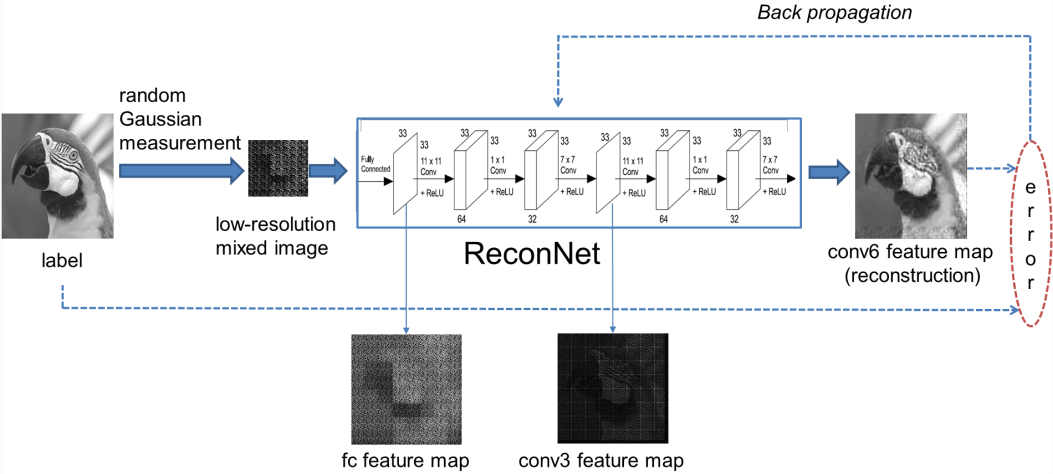}
	\caption{Framework of random Gaussian based network.}
	\label{fig:random_framework}
\end{figure}

Recently, deep learning methods have been applied in CS image reconstruction tasks and achieve promising results such as \cite{mousavi2015deep,kulkarni2016reconnet,mousavi2017deepcodec}. Among them, CNN-based methods present superior performance. ReconNet \cite{kulkarni2016reconnet} is a representative work that applies CNNs in reconstructing low-resolution mixed image measured by random Gaussian matrix. The framework is shown in Fig. \ref{fig:random_framework}.

The training of the network is driven by the error between the label and the reconstructed image. And the loss function is given by
\begin{eqnarray}\label{equation1}
L(\{W\})=\frac{1}{T}\sum_{i=1}^{T}{\lVert f(y_i,\{W\})-x_i \rVert}^2,
\end{eqnarray}
where $f(y_i,\{W\})$ is the $i-th$ reconstructed image of ReconNet. $x_i$ is the $i-th$ original signal as well as the $i-th$ label. $\{W\}$ means the training parameters in ReconNet. $T$ is the total number of image blocks in the training dataset. The loss function is minimized by tuning $\{W\}$ using back propagation.

Based on the way the original image is measured, deep learning methods for CS reconstruction can be divided into two categories: \emph{fixed random Gaussian measurement} and \emph{adaptive measurement}.
\paragraph{Fixed random Gaussian measurement}
Mousavi \emph{et al.} \cite{mousavi2015deep} firstly use SDA to recover signals from undersampled measurements. ReconNet \cite{kulkarni2016reconnet} and DeepInverse \cite{mousavi2017learning} utilizes CNNs to recover signals from Gaussian measurements. DR$^2$-Net \cite{yao2017dr}, inheriting ReconNet, adds residual connection blocks (Resblock) to its reconstruction stage and achieves better performance. Instead of learning to directly reconstruct the high-resolution image from the low-resolution one, DR$^2$-Net learns the residual between the ground truth image and the preliminary reconstructed image. However, the measurements of these methods are randomly measured, which is not optimally designed for natural images.

\paragraph{Adaptive measurement}
In order to keep the information of the training data, the adaptive measurement is proposed. Methods including improved ReconNet \cite{lohit2017convolutional}, Adp-Rec\footnote{Adp-Rec stands for adaptive measurement network for CS image reconstruction, proposed in our previous work.} \cite{xie2017adaptive}, and DeepCodec \cite{mousavi2017deepcodec} are all based on adaptive measurement. In cases of the improved ReconNet and Adp-Rec, a fully-connected layer is used to measure the signals, which allows for a jointly learning of the measurement and reconstruction stages. With the learned measurement matrix, a significant gain in terms of PSNR is achieved. DeepCodec, closely related to the DeepInverse, learns a transformation for dimensionality reduction. Learning measurements from the original signals helps to preserve more information compared with taking random measurements.

\section{Fully Convolutional Measurement}\label{section3}
\begin{figure}[h!]
	\centering
	\includegraphics[width=\linewidth]{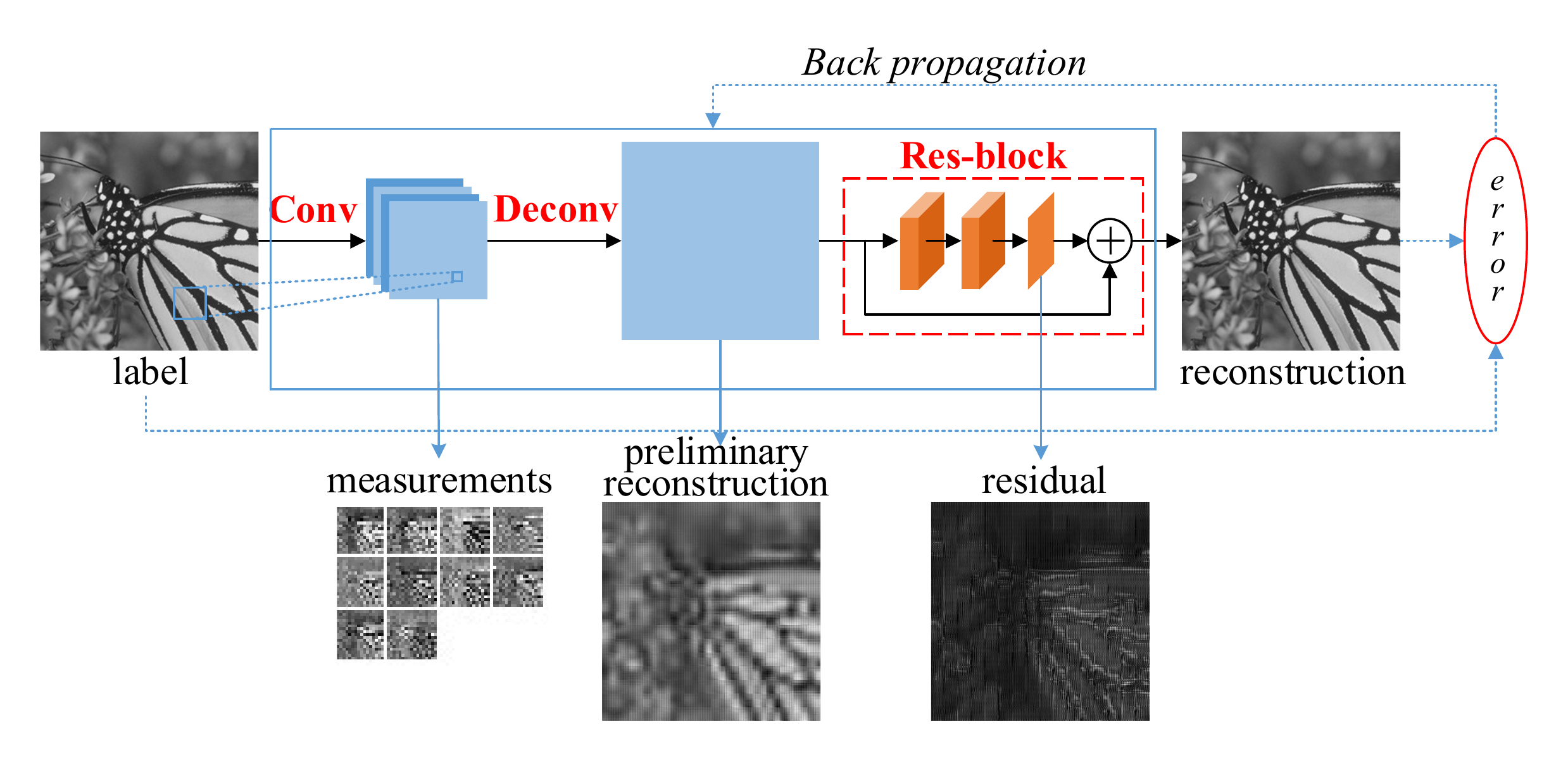}
	\caption{Framework of the proposed network.}
	\label{fig:framework}
\end{figure}

The exsiting CNN-based CS methods always adopt block-wise pattern due to the limitation of GPU memory. The block effect comes accordingly. In order to overcome this shortcoming, we propose a fully convolutional measurement network in which a convolutional layer is used to get the adaptive measurements. It is different from our previous work using fully-connected layers \cite{xie2017adaptive}. Fig. \ref{fig:framework} shows the proposed network  which is composed of a convolutional layer, a deconvolutional layer \cite{xu2014deep}, and a residual block. The first layer `conv' is used to obtain measurements. The second layer `deconv' is used to recover a low resolution image from the measurements. Furthermore, we apply a residual network (ResNet) to reconstruct the high resolution image. Because batch normalization would get rid of range flexibility from networks \cite{lim2017enhanced}, we remove the batch normalization layer in Resblock. Our framework is different from super-resolution (SR)~\cite{deng2016similarity}~\cite{fan2017compressed}~\cite{dong2016image}~\cite{Ledig_2017_CVPR}, since SR just learns a transform form the low-resolution images to high-resolution images, while the proposed framework is jointly trained from the measurement to the recovery part.

The loss function of the proposed network is given by
\begin{eqnarray}\label{equation2}
L(\{W\})=\frac{1}{T}\sum_{i=1}^{T}{\lVert f(x_i,\{W,K\})-x_i \rVert}^2,
\end{eqnarray}
where $K$ is the parameter of the convolutional measurement network, and $W$ is the parameters of the reconstruction network. The Euclidian distance between the label and the reconstruction is back propagated to train the whole network.

The main advantage of the proposed network is the use of convolutional layer as the measurement matrix. By means of the overlapped convolutional kernels, this structure can remove block effect of the reconstructed images. In details, one feature map contains several measurements of each pixel, which is similar to the idea proposed by He \emph{et al.} \cite{he2017mask} that the feature map preserves the explicit per-pixel spatial correspondence. Another advantage is that the fully convolutional neural network can deal with images of any size, which breaks the limitation that fully-connected layer is only capable of measuring the fixed size of images. Once the network is trained, we can test images with different sizes without changing the structure of the network.

\begin{figure}[h!]
	\centering
	\subfloat[Origin]{\includegraphics[width=0.2\linewidth]{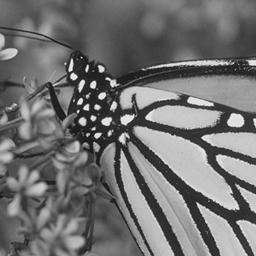}} \qquad
	\subfloat[ ReconNet \protect \\ \centerline{ (21.49dB)}]{\includegraphics[width=0.2\linewidth]{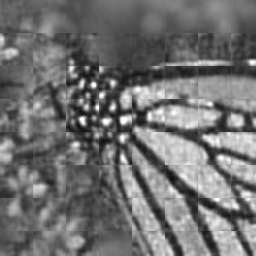}} \qquad
	\subfloat[\quad Adp-Rec \protect \\ \centerline{\quad (26.65dB)}]{\includegraphics[width=0.2\linewidth]{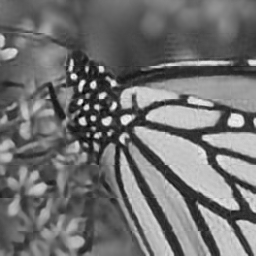}} \qquad
	\subfloat[\color{red} { Proposed \protect \\ \centerline{ (27.61dB)}}]{\includegraphics[width=0.2\linewidth]{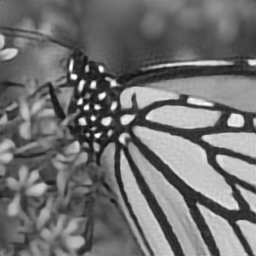}}
	\caption{The reconstruction results of monarch at measurement rate 10\%.}
	\label{fig:Monarch}
\end{figure}

Fig. \ref{fig:Monarch} shows an example of reconstruction results at three kinds of measurements. The original image and those by random Gaussian, Adp-Rec, and the proposed method are shown respectively in Fig. \ref{fig:Monarch}(a), (b), (c), and (d). The measurement rate is 10\%. We can see that the proposed method performs better than the others.

\begin{figure}[h!]
	\centering
	\subfloat[Random Gaussian measurements.]{\includegraphics[width=\linewidth]{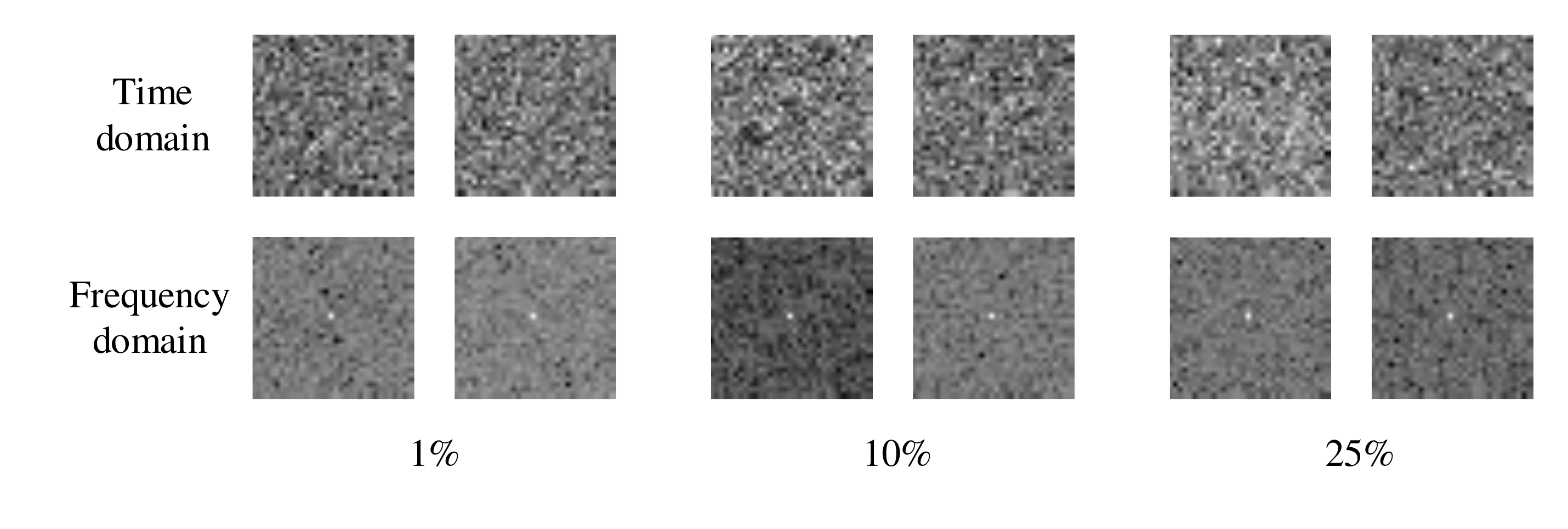}} \\
	\subfloat[Adaptive measurements in Adp-Rec.]{\includegraphics[width=\linewidth]{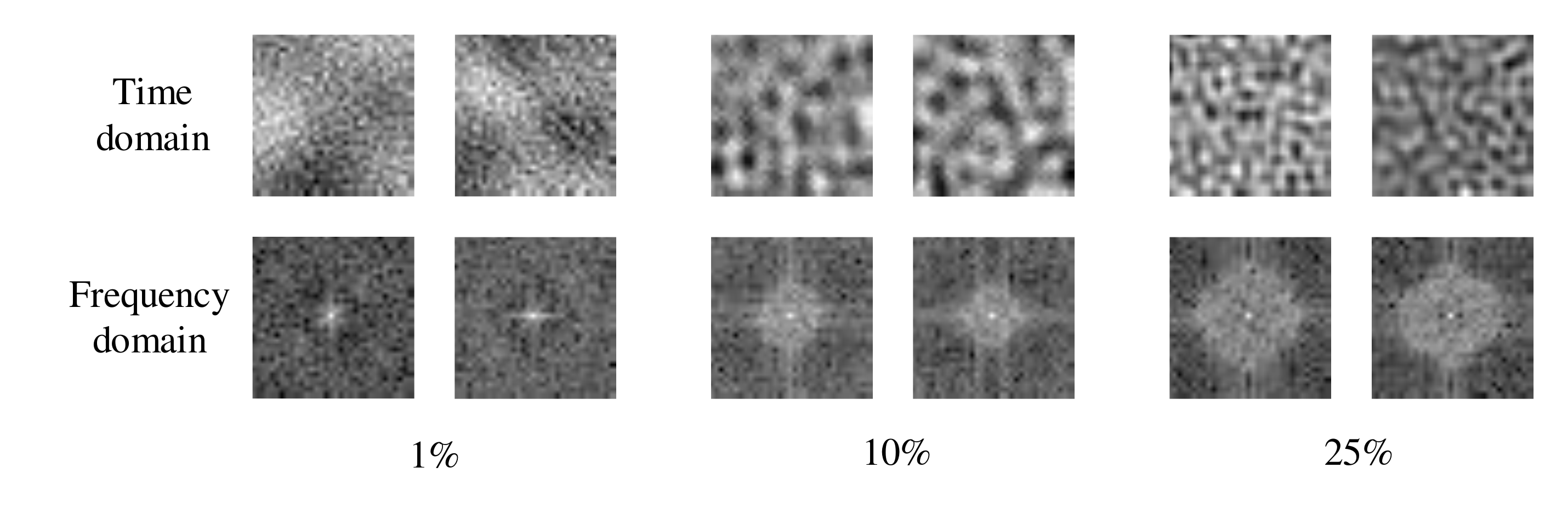}} \\
	\subfloat[Proposed]{\includegraphics[width=\linewidth]{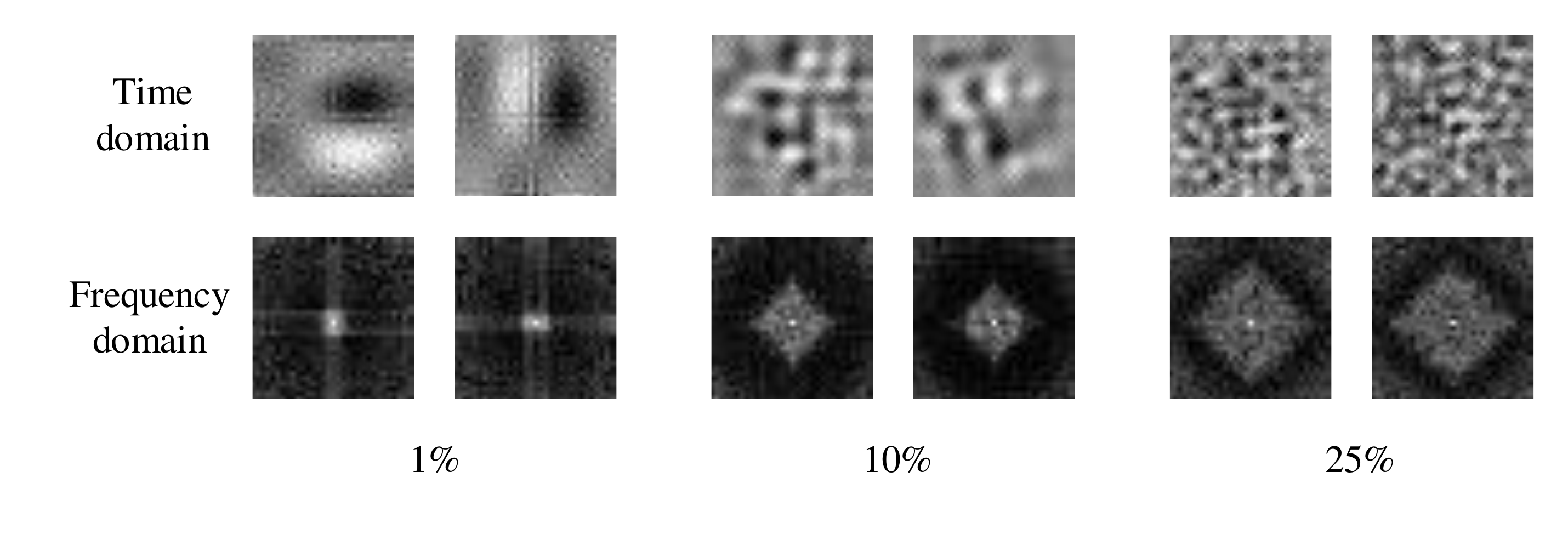}}
	\caption{Reshaped row vectors of measurement matrix at measurement rates 1\%, 10\%, and 25\% in both time and frequency domain.}
	\label{fig:measurements}
\end{figure}

The better performance can be proved through a visualization of the kernels in convolutional layer of the measurement network, as shown in Fig. \ref{fig:measurements}. Since the original signal in  random Gaussian and adaptive measurements is a cloumn vector (Fig. \ref{fig:measurements}(a) and (b)), we reshape the row vectors of measurement matrix to size $33\times33$. Fig. \ref{fig:measurements}(a) shows two reshaped row vectors of the random Gaussian measurement matrix at measurement rates 1\% , 10\%, and 25\% in both time and frequency domain. The content of random Gaussian measurement matrix is obviously irregular. Fig. \ref{fig:measurements}(b) shows two reshaped row vectors of adaptive measurement matrix in Adp-Rec. When measurement rate is set to 1\%, low frequency information is already extracted. As the measurement rate increases, much high frequency information is  captured. Fig. \ref{fig:measurements}(c) shows two kernels of the proposed measurement matrix. Compared with the adaptive measurements in Adp-Rec, the measurements by the proposed method provide more concentrated energy in the low frequency area at different measurement rates. As for the directional information, when measurement rate is 1\%, two extracted typical directions `horizontal' and `vertical' can be easily observed in time domain.

\begin{figure}[h!]
	\centering
	\includegraphics[width=\linewidth]{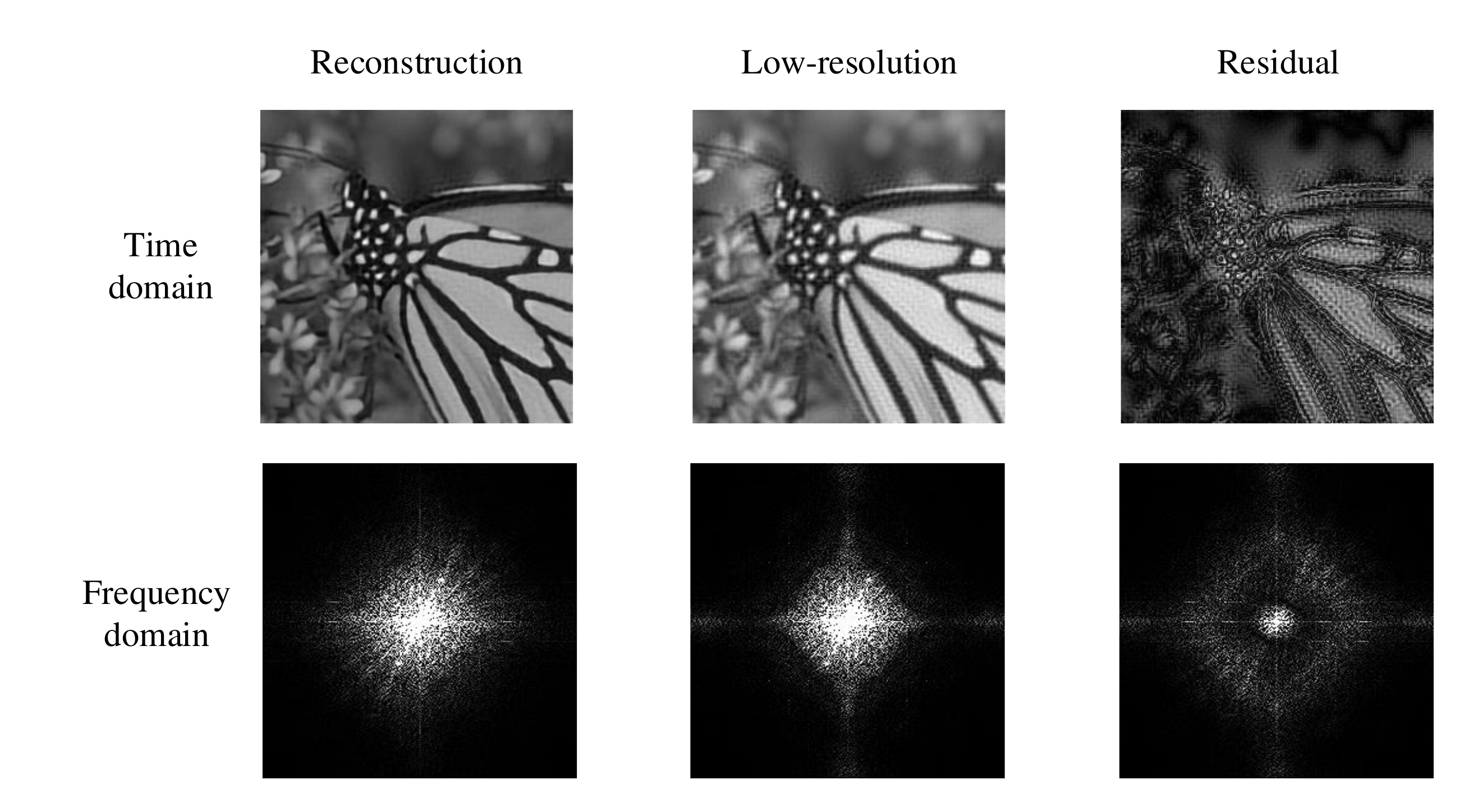}
	\caption{Reconstruction image, low-resolution image and residual image at measurement rate 10\%  in both time and frequency domain.}
	\label{fig:Monarch_res10}
\end{figure}

Fig. \ref{fig:Monarch_res10} shows the reconstruction of image `Monarch', its low-resolution, and the corresponding residual.
From residual image in frequency domain, we can see that the high frequency component is mainly learned by the residual network.  Rather than ReconNet which reconstructs the high resolution image from the low resolution one directly, ResNet just reconstruct the residual between the low resolution image and the high resolution image, that is the reconstruction image. Thus, all its energy is concentrated on the residual. That is why ResNet has better performance.

\section{Experiments}\label{section4}
In this section, we perform experiments on the reconstruction of compressive sensing images with existing typical methods. The results show the outstanding performance by the proposed method.

The experiments are conducted on caffe framework \cite{jia2014caffe}. Our computer is equipped with Intel Core i7-6700 CPU with frequency of 3.4GHz, 4 NVidia GeForce GTX Titan XP GPUs, 128 GB RAM, and the framework runs on Ubuntu 16.04 operating system. The training dataset consists of 800 pieces of $256\times256$ size images downsampled and divided from 800 images in DIV2K dataset \cite{timofte2017ntire}.

\begin{figure}[h!]
	\centering
	\subfloat[Origin]{\includegraphics[width=0.2\linewidth]{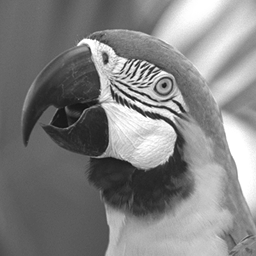}} \qquad
	\subfloat[ ReconNet \protect \\ \centerline{ (18.93dB)}]{\includegraphics[width=0.2\linewidth]{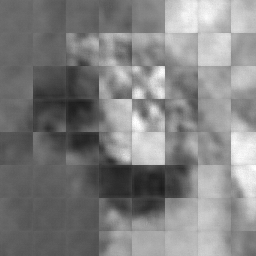}} \qquad
	\subfloat[\quad Adp-Rec \protect \\ \centerline{\quad (21.67dB)}]{\includegraphics[width=0.2\linewidth]{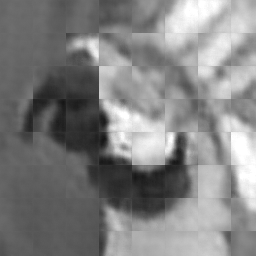}} \qquad
	\subfloat[{\color{red} { Proposed \protect \\ \centerline{ (22.49dB)}}}]{\includegraphics[width=0.2\linewidth]{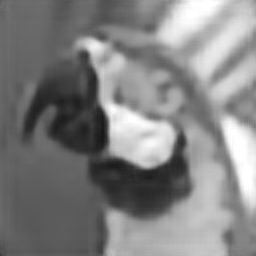}}
	\caption{The reconstruction results of parrots at measurement rate 1\%.}
	\label{fig:Parrots}
\end{figure}

\begin{figure}[h!]
	\centering
	\subfloat[Origin]{\includegraphics[width=0.2\linewidth]{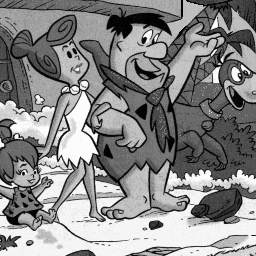}} \qquad
	\subfloat[ ReconNet \protect \\ \centerline{ (19.04dB)}]{\includegraphics[width=0.2\linewidth]{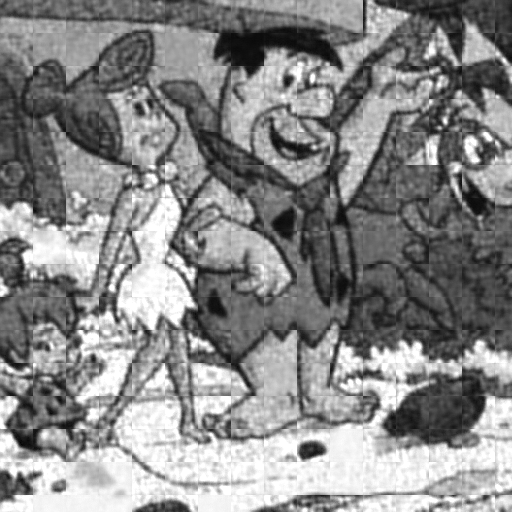}} \qquad
	\subfloat[\quad Adp-Rec \protect \\ \centerline{\quad (23.83dB)}]{\includegraphics[width=0.2\linewidth]{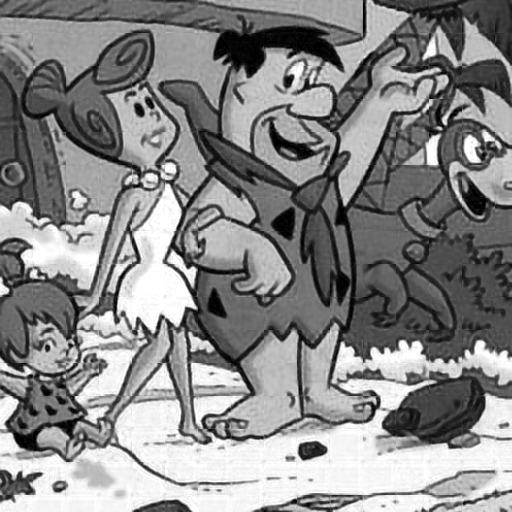}} \qquad
	\subfloat[{\color{red} { Proposed \protect \\ \centerline{ (24.98dB)}}}]{\includegraphics[width=0.2\linewidth]{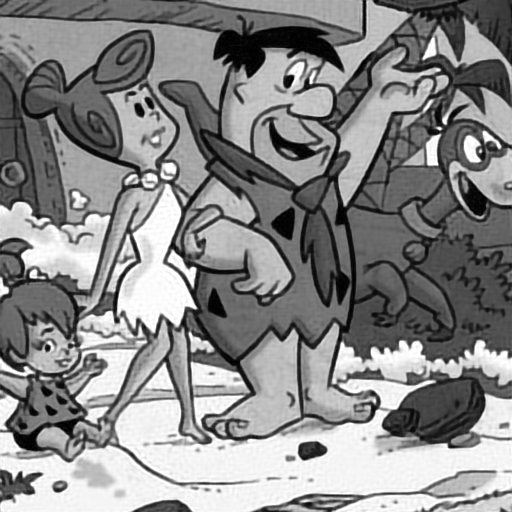}}
	\caption{The reconstruction results of flinstones at measurement rate 10\%.}
	\label{fig:Flinstones}
\end{figure}

\begin{figure}[h!]
	\centering
	\subfloat[Origin]{\includegraphics[width=0.2\linewidth]{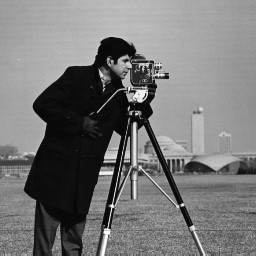}} \qquad
	\subfloat[ ReconNet \protect \\ \centerline{ (23.48dB)}]{\includegraphics[width=0.2\linewidth]{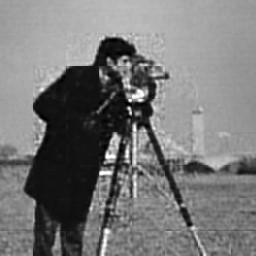}} \qquad
	\subfloat[\quad Adp-Rec \protect \\ \centerline{\quad (27.11dB)}]{\includegraphics[width=0.2\linewidth]{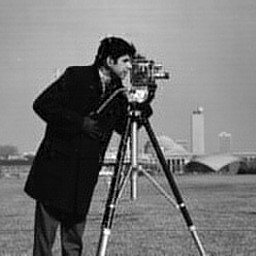}} \qquad
	\subfloat[{\color{red} { Proposed \protect \\ \centerline{ (28.99dB)}}}]{\includegraphics[width=0.2\linewidth]{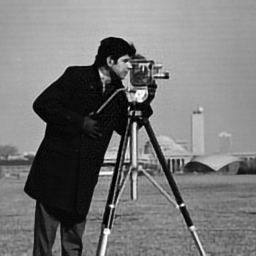}}
	\caption{The reconstruction results of cameraman at measurement rate 25\%.}
	\label{fig:Cameraman}
\end{figure}

The performance of the proposed method is compared with those by ReconNet and Adp-Rec which are the typical CNN-based CS methods. We give the testing results using image `parrots', `flinstones', and `cameraman' at measurement rates 1\%, 10\%, and 25\%, as shown in Fig. \ref{fig:Parrots}, Fig. \ref{fig:Flinstones}, and Fig. \ref{fig:Cameraman}, respectively. The proposed method provides the best reconstruction results in terms of PSNR and the results are most visually attractive.

From the results shown in Fig. \ref{fig:Parrots}, with measurement rate being 1\%, it can be seen that the block effect is removed (Fig \ref{fig:Parrots}(d) vs. (b) and (c)). From Fig. \ref{fig:Flinstones}, when the measurement rate is 10\%, the proposed method shows the advantage in reconstructing the image, typically in those smooth areas such as nose, hands, and legs of the man. From Fig. \ref{fig:Cameraman}, when measurement rate rises to 25\%, the proposed method still outperforms other methods, which can be easily seen in the edge of the man's arm.

\newpage
For an overall look on the performance, the reconstruction results of 11 test images at measurement rates 1\%, 10\%, and 25\% with the methods including ReconNet, DR$^2$-Net, Adp-Rec, Fully-Conv\footnote{Fully-Conv consists of a convolutional layer and a deconvolutional layer without Resblock, which can be regarded as the tiny model of the proposed network.}, and the proposed one are shown in Table \ref{tab:PSNR}. The mean PSNR is given in the type of blue background. It is obvious that the proposed method shows greatest performance in almost all test images.

\begin{table}[h!]
	\centering
	\footnotesize
	\caption{The PSNR results at measurement rates (MR) 1\%, 10\%, and 25\%, where Red is ranked the first and blue is ranked the second.}
	\label{tab:PSNR}
	\renewcommand\arraystretch{0.8}
	\begin{tabular}{|p{0.5cm}|p{1.7cm}ccccc|}
		\hline
		\rowcolor[HTML]{87B0FF}
		\textbf{MR} & \textbf{Samples}           & \textbf{ReconNet} & \textbf{$\bf{DR^2}$-Net} & \textbf{Adp-Rec}               & \textbf{Fully-Conv}            & \textbf{Proposed}              \\ \hline
		\multirow{12}{0.5cm}{1\%} & \textbf{Monarch}          & 15.61dB           & 15.33dB                       & {17.70dB} & {\color[HTML]{0000FF} 17.98dB} & {\color[HTML]{CC0000} 18.46dB} \\
		& \textbf{Parrots}          & 18.93dB           & 18.01dB                       & {21.67dB} & {\color[HTML]{0000FF} 21.80dB} & {\color[HTML]{CC0000} 22.49dB} \\
		& \textbf{Barbara}          & 19.08dB           & 18.65dB                       & {21.36dB} & {\color[HTML]{0000FF} 21.61dB} & {\color[HTML]{CC0000} 22.06dB} \\
		& \textbf{Boats}            & 18.82dB           & 18.67dB                       & {21.09dB} & {\color[HTML]{0000FF} 21.73dB} & {\color[HTML]{CC0000} 22.3dB}  \\
		& \textbf{Cameraman}        & 17.51dB           & 17.08dB                       & {19.74dB} & {\color[HTML]{0000FF} 19.88dB} & {\color[HTML]{CC0000} 20.63dB} \\
		& \textbf{Fingerprint} & 15.01dB           & 14.73dB                       & {16.22dB} & {\color[HTML]{0000FF} 16.24dB} & {\color[HTML]{CC0000} 16.33dB} \\
		& \textbf{Flinstones}  & 14.14dB           & 14.01dB                       & {16.12dB} & {\color[HTML]{0000FF} 16.55dB} & {\color[HTML]{CC0000} 16.92dB} \\
		& \textbf{Foreman}          & 22.03dB           & 20.59dB                       & {\color[HTML]{0000FF} 25.53dB} & {25.18dB} & {\color[HTML]{CC0000} 27.26dB} \\
		& \textbf{House}            & 20.30dB           & 19.61dB                       & {\color[HTML]{0000FF} 22.93dB} & {\color[HTML]{0000FF} 22.93dB} & {\color[HTML]{CC0000} 23.67dB} \\
		& \textbf{Lena}             & 18.51dB           & 17.97dB                       & {21.49dB} & {\color[HTML]{0000FF} 21.77dB} & {\color[HTML]{CC0000} 22.51dB} \\
		& \textbf{Peppers}          & 17.39dB           & 16.90dB                       & {19.75dB} & {\color[HTML]{0000FF} 20.80dB} & {\color[HTML]{CC0000} 21.38dB} \\
		& \cellcolor[HTML]{BCCCFF}\textbf{Mean(all)}        & \cellcolor[HTML]{BCCCFF}17.94dB           & \cellcolor[HTML]{BCCCFF}17.44dB                       & \cellcolor[HTML]{BCCCFF}{20.33dB} & \cellcolor[HTML]{BCCCFF}{\color[HTML]{0000FF} 20.59dB} & \cellcolor[HTML]{BCCCFF}{\color[HTML]{CC0000} 21.27dB} \\
		\hline
		\multirow{12}{0.5cm}{10\%} & \textbf{Monarch}     & 21.49dB           & 23.10dB          & {\color[HTML]{0000FF} 26.65dB} & {25.20dB} & {\color[HTML]{CC0000} 27.61dB} \\
		& \textbf{Parrots}     & 23.36dB           & 23.94dB          & {\color[HTML]{0000FF} 27.59dB} & {26.82dB} & {\color[HTML]{CC0000} 27.92dB} \\
		& \textbf{Barbara}     & 22.17dB           & 22.69dB          & {\color[HTML]{0000FF} 24.28dB} & {\color[HTML]{CC0000} 24.39dB} & {\color[HTML]{0000FF} 24.28dB} \\
		& \textbf{Boats}       & 24.56dB           & 25.58dB          & {\color[HTML]{0000FF} 28.80dB} & {28.52dB} & {\color[HTML]{CC0000} 29.48dB} \\
		& \textbf{Cameraman}   & 21.54dB           & 22.46dB          & {\color[HTML]{0000FF} 24.97dB} & {24.58dB} & {\color[HTML]{CC0000} 25.62dB} \\
		& \textbf{Fingerprint} & 20.99dB           & 22.03dB          & {26.55dB} & {\color[HTML]{0000FF} 26.92dB} & {\color[HTML]{CC0000} 27.36dB} \\
		& \textbf{Flinstones}  & 19.04dB           & 21.09dB          & {\color[HTML]{0000FF} 23.83dB} & {23.08dB} & {\color[HTML]{CC0000} 24.98dB} \\
		& \textbf{Foreman}     & 29.02dB           & 29.20dB          & {\color[HTML]{0000FF} 33.51dB} & {31.96dB} & {\color[HTML]{CC0000} 34.00dB} \\
		& \textbf{House}       & 26.74dB           & 27.53dB          & {\color[HTML]{0000FF} 31.43dB} & {30.81dB} & {\color[HTML]{CC0000} 32.36dB} \\
		& \textbf{Lena}        & 24.48dB           & 25.39dB          & {\color[HTML]{0000FF} 28.50dB} & {27.76dB} & {\color[HTML]{CC0000} 28.97dB} \\
		& \textbf{Peppers}     & 22.72dB           & 24.32dB          & {26.67dB} & {\color[HTML]{0000FF} 26.69dB} & {\color[HTML]{CC0000} 28.72dB} \\
		& \cellcolor[HTML]{BCCCFF}\textbf{Mean(all)}   & \cellcolor[HTML]{BCCCFF}23.28dB           & \cellcolor[HTML]{BCCCFF}24.32dB          & \cellcolor[HTML]{BCCCFF}{\color[HTML]{0000FF} 27.53dB} & \cellcolor[HTML]{BCCCFF}{26.98dB} & \cellcolor[HTML]{BCCCFF}{\color[HTML]{CC0000} 28.30dB} \\
		\hline
		\multirow{12}{0.5cm}{25\%} & \textbf{Monarch}     & 24.95dB           & 27.95dB          & {\color[HTML]{0000FF} 29.25dB} & {28.47dB} & {\color[HTML]{CC0000} 32.63dB} \\
		& \textbf{Parrots}     & 26.66dB           & 28.73dB          & {\color[HTML]{0000FF} 30.51dB} & {29.90dB} & {\color[HTML]{CC0000} 32.13dB} \\
		& \textbf{Barbara}     & 23.58dB           & 25.77dB          & {\color[HTML]{0000FF} 27.40dB} & {27.11dB} & {\color[HTML]{CC0000} 28.59dB} \\
		& \textbf{Boats}       & 27.83dB           & 30.09dB          & {\color[HTML]{0000FF} 32.47dB} & {31.75dB} & {\color[HTML]{CC0000} 33.88dB} \\
		& \textbf{Cameraman}   & 23.48dB           & 25.62dB          & {\color[HTML]{0000FF} 27.11dB} & {26.73dB} & {\color[HTML]{CC0000} 28.99dB} \\
		& \textbf{Fingerprint} & 26.15dB           & 27.65dB          & {\color[HTML]{0000FF} 32.31dB} & {30.92dB} & {\color[HTML]{CC0000} 32.91dB} \\
		& \textbf{Flinstones}  & 22.74dB           & 26.19dB          & {\color[HTML]{0000FF} 27.94dB} & {27.02dB} & {\color[HTML]{CC0000} 30.26dB} \\
		& \textbf{Foreman}     & 32.08dB           & 33.53dB          & {\color[HTML]{0000FF} 36.18dB} & {35.08dB} & {\color[HTML]{CC0000} 38.10dB} \\
		& \textbf{House}       & 29.96dB           & 31.83dB          & {\color[HTML]{0000FF} 34.38dB} & {33.63dB} & {\color[HTML]{CC0000} 36.22dB} \\
		& \textbf{Lena}        & 27.47dB           & 29.42dB          & {\color[HTML]{0000FF} 31.63dB} & {30.65dB} & {\color[HTML]{CC0000} 33.00dB} \\
		& \textbf{Peppers}     & 25.74dB           & 28.49dB          & {29.65dB} & {\color[HTML]{0000FF} 29.71dB} & {\color[HTML]{CC0000} 32.90dB} \\
		& \cellcolor[HTML]{BCCCFF}\textbf{Mean(all)}   & \cellcolor[HTML]{BCCCFF}26.42dB           & \cellcolor[HTML]{BCCCFF}28.66dB          & \cellcolor[HTML]{BCCCFF}{30.80dB} & \cellcolor[HTML]{BCCCFF}{\color[HTML]{0000FF} 30.09dB} & \cellcolor[HTML]{BCCCFF}{\color[HTML]{CC0000} 32.69dB} \\
		\hline
	\end{tabular}
\end{table}

\begin{table}[h!]
	\centering
	\footnotesize
	\caption{The SSIM and MOS results. Here measurement rates (MR) 1\% is taken as an example. The highest is marked red, while the second is marked blue.
%Red is ranked the first and blue is ranked the second except for the original image.
}
	\label{tab:SSIM}
	\renewcommand\arraystretch{0.8}
	\begin{tabular}{|c|cccccc|}
		\hline
		\rowcolor[HTML]{87B0FF}
		& \textbf{Samples} & \textbf{Original} & \textbf{ReconNet} & \textbf{$\bf{DR^2}$-Net} & \textbf{Adp-Rec}  & \textbf{Proposed}   \\
\multirow{12}{0.8cm}{\textbf{MOS}} & \textbf{Monarch}          & 4.9615           & 1.0000                       & {1.1538} & {\color[HTML]{0000FF} 1.7307} & {\color[HTML]{CC0000} 2.4615} \\
		& \textbf{Parrots}        & 4.9615           & 1.0384                       & {1.2307} & {\color[HTML]{0000FF} 2.1538} & {\color[HTML]{CC0000} 2.9230} \\
		& \textbf{Barbara}        & 4.9615           & 1.0769                       & {1.0769} & {\color[HTML]{0000FF} 2.0000} & {\color[HTML]{CC0000} 2.6538} \\
		& \textbf{Boats}          & 4.9230           & 1.0769                       & {1.0384} & {\color[HTML]{0000FF} 1.5000} & {\color[HTML]{CC0000} 2.3846}  \\
		& \textbf{Cameraman}      & 5.0000           & 1.1538                       & {1.1923} & {\color[HTML]{0000FF} 1.8461} & {\color[HTML]{CC0000} 2.7692} \\
		& \textbf{Fingerprint}    & 4.8461           & 1.1538                       & {1.0384} & {\color[HTML]{0000FF} 1.4230} & {\color[HTML]{CC0000} 1.6823} \\
		& \textbf{Flinstones}     & 5.0000           & 1.1923                       & {1.1538} & {\color[HTML]{0000FF} 2.0769} & {\color[HTML]{CC0000} 3.1538} \\
		& \textbf{Foreman}        & 4.9230           & 1.1538                       & {1.1538} & {\color[HTML]{0000FF} 1.9230} & {\color[HTML]{CC0000} 2.7692} \\
		& \textbf{House}          & 4.9615           & 1.0000                       & {1.1153} & {\color[HTML]{0000FF} 2.0769} & {\color[HTML]{CC0000} 2.7307} \\
		& \textbf{Lena}           & 5.0000           & 1.0384                       & {1.0384} & {\color[HTML]{0000FF} 1.8076} & {\color[HTML]{CC0000} 2.8461} \\
		& \textbf{Peppers}        & 4.9615           & 1.0000                       & {1.1153} & {\color[HTML]{0000FF} 1.8076} & {\color[HTML]{CC0000} 2.5769} \\
		& \cellcolor[HTML]{BCCCFF}\textbf{Mean(all)} & \cellcolor[HTML]{BCCCFF}4.9545  & \cellcolor[HTML]{BCCCFF}1.0734  & \cellcolor[HTML]{BCCCFF}1.1188 & \cellcolor[HTML]{BCCCFF}{\color[HTML]{0000FF} 1.8496} & \cellcolor[HTML]{BCCCFF}{\color[HTML]{CC0000} 2.6328} \\
		\hline
\multirow{12}{0.8cm}{\textbf{SSIM}} & \textbf{Monarch}          & 1.0000           & 0.3801                       & {0.3931} & {\color[HTML]{0000FF} 0.4755} & {\color[HTML]{CC0000} 0.5058} \\
		& \textbf{Parrots}        & 1.0000           & 0.5328                       & {0.5617} & {\color[HTML]{0000FF} 0.6739} & {\color[HTML]{CC0000} 0.7135} \\
		& \textbf{Barbara}        & 1.0000           & 0.3729                       & {0.3847} & {\color[HTML]{0000FF} 0.4648} & {\color[HTML]{CC0000} 0.5007} \\
		& \textbf{Boats}          & 1.0000           & 0.4140                       & {0.4319} & {\color[HTML]{0000FF} 0.4888} & {\color[HTML]{CC0000} 0.5405 }  \\
		& \textbf{Cameraman}      & 1.0000           & 0.4516                       & {0.4783} & {\color[HTML]{0000FF} 0.5578} & {\color[HTML]{CC0000} 0.5867} \\
		& \textbf{Fingerprint}    & 1.0000           & 0.1548                       & {\color[HTML]{CC0000} 0.1727} & {0.1628} & {\color[HTML]{0000FF}0.1700} \\
		& \textbf{Flinstones}     & 1.0000           & 0.2502                       & {0.2718} & {\color[HTML]{0000FF} 0.3230} & {\color[HTML]{CC0000} 0.3801} \\
		& \textbf{Foreman}        & 1.0000           & 0.5647                       & {0.6051} & {\color[HTML]{0000FF} 0.6912} & {\color[HTML]{CC0000} 0.7396} \\
		& \textbf{House}          & 1.0000           & 0.5278                       & {0.5526} & {\color[HTML]{0000FF} 0.6350} & {\color[HTML]{CC0000} 0.6624} \\
		& \textbf{Lena}           & 1.0000           & 0.4418                       & {0.4552} & {\color[HTML]{0000FF} 0.5554} & {\color[HTML]{CC0000} 0.6081} \\
		& \textbf{Peppers}        & 1.0000           & 0.4002                       & {0.4127} & {\color[HTML]{0000FF} 0.5053} & {\color[HTML]{CC0000} 0.5839} \\
		& \cellcolor[HTML]{BCCCFF}\textbf{Mean(all)} & \cellcolor[HTML]{BCCCFF}1.0000  & \cellcolor[HTML]{BCCCFF}0.4083  & \cellcolor[HTML]{BCCCFF}0.4291 & \cellcolor[HTML]{BCCCFF}{\color[HTML]{0000FF} 0.5031} & \cellcolor[HTML]{BCCCFF}{\color[HTML]{CC0000} 0.5447} \\
		\hline
		%\textbf{SSIM}   & 1  & 0.4083  & 0.4291 & {\color[HTML]{0000FF} 0.5031} & {\color[HTML]{CC0000} 0.5447} \\
		%\hline
		%\textbf{MOS} & 4.9545 & 1.0734 & 1.1189 & {\color[HTML]{0000FF} 1.8497} & {\color[HTML]{CC0000} 2.6329} \\
		%\hline
	\end{tabular}
\end{table}

From Table \ref{tab:PSNR}, it can be concluded that Adp-Rec beats DR$^2$-Net and ReconNet about 3dB in all measurement rates because of its adaptive measurement. Based on the standard ReconNet \cite{kulkarni2016reconnet}, the improved ReconNet \cite{lohit2017convolutional} adds several tricks such as adaptive measurement and adversarial loss. Its performance is even lower than Adp-Rec. Despite its promising results, Adp-Rec still divides image into blocks, ignoring the relevance between neighbouring blocks, which causes to the block effect in reconstructed images. For this reason, Fully-Conv uses a convolution layer as measurement matrix to deal with this problem. It achieves comparable results with Adp-Rec even though it contains no additional operation.

To further improve the reconstruction results, we put Resblock after Fully-Conv structure because of the brilliant performance of Resblock in reconstruction task. With this enhancement, the proposed method obtains the best performance in terms of PSNR at all measurement rates, as shown in Table \ref{tab:PSNR}.

We also measure the quality of images with Mean Opinion Score (MOS). The test results of different images are shown in Table \ref{tab:SSIM}. In this experiment, 26 volunteers take part in ranking the images. The quality of the images is divided into five levels, from 1 to 5, with the quality from low to high. All the test images are randomly ranked before being scored and they are displayed group by group. Each group has four reconstruction images, in different methods, and one original scene image. All participants take this test on the same computer screen, from the same angle and distance. Here the distance from the screen to the tested persons is 50 cm and the eyes of those persons are of the same height of the center of the screen. In addition, we also use structural similarity index (SSIM) to evaluate our method and existing block-wised methods as shown in Table \ref{tab:SSIM}. The case of MR = 1\% is taken as an example.

In terms of hardware implementation, we follow the approach of the previous work proposed in \cite{shi2011high} in which sliding window is used to measure the scene. Similarly, we can replace the random Gaussian measurement matrix with the learned pre-defined  parameters in the conv layer of the measurement network. The reconstruction part is not on optical device, so only the measurement part needs to be implemented with the approach above.

\section{Conclusion}\label{Conclusion}\label{conclusion}
This paper proposes a novel CNN-based deep neural network for high-quality compressive sensing image reconstruction. The network uses a fully convolutional architecture, which removes the block effect caused by block-wise methods. For a further improvement, we add Resblock after the deconvolutional layer, making the network learn the residual information between low and high resolution images. With this enhancement, the network shows best performance in reconstruction task compared with other methods. In future work, we are going to apply perceptual loss into the network for better reconstruction result. And semantics-oriented reconstruction will be also considered.

\bibliographystyle{elsarticle-num}
\bibliography{Reference}
\end{document}